# Kernel Bandwidth Selection for SVDD:

# The Sampling Peak Criterion Method for Large Data

Sergiy Peredriy, Deovrat Kakde, and Arin Chaudhuri

SAS Institute Inc.

Cary, NC, USA

**Abstract:** Support vector data description (SVDD) provides a useful approach, with various practical applications, for constructing a description of multivariate data for single-class classification and outlier detection. The Gaussian kernel that is used in SVDD formulation allows a flexible data description defined by observations that are designated as support vectors. The data boundary of such a description is nonspherical and conforms to the geometric features of the data. By varying the Gaussian kernel bandwidth parameter, the SVDD-generated boundary can be made either smoother (more spherical, which might lead to underfitting), or tighter and more jagged (which might result in overfitting). Kakde et al. [1] proposed a peak criterion for selecting an optimal value of the kernel bandwidth to strike a balance between the data boundary smoothness and a method's ability to capture the general geometric shape of the data. The peak criterion approach involves training the SVDD at various values of the kernel bandwidth parameter. When training data sets are large, the time required to obtain the optimal value of the Gaussian kernel bandwidth parameter according to the peak method can become prohibitively large. This paper proposes an extension of the peak method for the case of large data. The proposed method produces good results when applied to several data sets. Two existing alternative methods of computing the Gaussian kernel bandwidth parameter (coefficient of variation and distance to the farthest neighbor) were modified in order to allow comparison with the proposed method on convergence. Empirical comparison demonstrates the advantage of the proposed method.

## 1. Introduction

Support vector data description (SVDD) is a machine-learning algorithm that is used to build a flexible description of data of a single class and to detect outliers. SVDD is an extension of support vector machines and was first introduced by Tax and Duin [7]. A data boundary is characterized by observations that are designated as support vectors. SVDD is applied in domains in which all or the majority of the data belongs to a single class. Among existing and proposed applications are multivariate process control [8], machine condition monitoring [9, 10], and image classification [11].

Mathematical formulation of the SVDD model is detailed in Appendix A. In the primal form, the SVDD model builds a minimum-radius hypersphere around the data. However, if the expected data distribution is not spherical, this method might lead to extra space being included within the boundary, leading to false positives when scoring. Therefore, it is desirable to have a compact boundary around the data that would conform to the geometric shape of the single-class data. This goal is possible with the use of kernel functions in the objective function (see Appendix A for the details). In this paper, the Gaussian kernel function is used,

$$K(x_i, x_j) = exp \frac{-\left\| x_i - x_j \right\|^2}{2s^2} \tag{1}$$

where $x_i \in \mathbb{R}^m, i = 1, \ldots, n$ is the training data and $s$ is the Gaussian bandwidth parameter.

## 1.1 Kernel Bandwidth Value and The Peak Criterion Method

Kakde et al. [1] points out that, if outlier fraction $f$ is kept constant, the behavior of the boundary around the data and the number of support vectors generated by the SVDD algorithm depend on the selected value of the Gaussian kernel bandwidth parameter $s$.[1] For a relatively small value of $s$, the boundary is very jagged and the number of support vectors is large, approaching the number of observations in the training data set. At extremely low values of $s$, every data point becomes a support vector. As the value of $s$ increases, the number of support vectors declines, and the boundary becomes smoother and starts following the shape of the data. However, as $s$ increases further, the boundary becomes overly smooth and it no longer conforms to the shape of the data. At extremely large values of $s$, the boundary becomes spherical and the number of support vectors reaches its minimum.

It is important to select a value of $s$ that creates a relatively tight boundary around training data while conforming to the general shape of the data and without being too jagged. Selecting a proper value of $s$ can be challenging. Kakde et al. [1] offer the peak method of selecting an optimal value of $s$. At the core of the peak method is the observation that near the value of $s$ that produces a good boundary around the training data, the second derivative (with respect to $s$) of the optimal objective function (Equation A.8) crosses 0, indicating the first extremum of the first derivative. The peak approach of Kakde et al. [1] can be summarized as follows:

1. Obtain an SVDD model by using the training data set for a range of values of $s$ large enough to include the optimal value, with a small step $\Delta s$, and obtain the value of the optimal objective function (OOF) for each value of $s$.
2. Approximate the first and second derivatives of OOF with the following differences, respectively:
$$\frac{df}{ds} \approx dif(s) = \frac{f(s+\Delta s) - f(s)}{\Delta s}$$
$$\frac{d^2 f}{ds^2} \approx \frac{dif(s+\Delta s) - dif(s)}{\Delta s} = \frac{f(s+2\Delta s) - 2f(s+\Delta s) + f(s)}{(\Delta s)^2}$$
3. Smooth the second difference of OOF with a penalized B-spline.
4. Choose the value of $s$ from the range where the 95% confidence interval of the smoothed second difference obtained in step 3 includes 0 for the first time.

Kakde et al. [1] demonstrate the viability of the method by using several two-dimensional data sets, as well as samples from large higher-dimensional data sets. This method is referred to as "full peak method" throughout this paper.

The SVDD approach requires solving a quadratic programming problem. The time needed to solve the quadratic programming problem is directly related to the size of the training data set. Moreover, we observed that the computation time is higher for smaller values of $s$.[2] With smaller values of $s$, the solution to the SVDD problem is more complex, resulting in a large number of support vectors and requiring more computational time. Figure 1 shows the computational time to solve the SVDD problem for different values of $s$ when using the full peak method for two large multidimensional data sets. Figure 1a is for the Shuttle data set, which has 45,586 observations and nine variables; Figure 1b is for the Tennessee Eastman (TE) data set, which has 108,000 observations and 41 variables.[3] In both cases, $s$ was changed from 0.5 to 60 with a step of 0.5. As can be seen from the graph, smaller values of $s$ require significant computational time (reaching 7:00 minutes for $s = 0.5$ for the TE data). Moreover, you can see that a larger data set requires more time to solve. The total time for this exercise for both data sets is summarized in Table 1. The reported times are totals from Figure 1.

---

[1] See Equation A.1 in the Appendix A.
[2] We used the SVM algorithm to solve the SVDD problem.

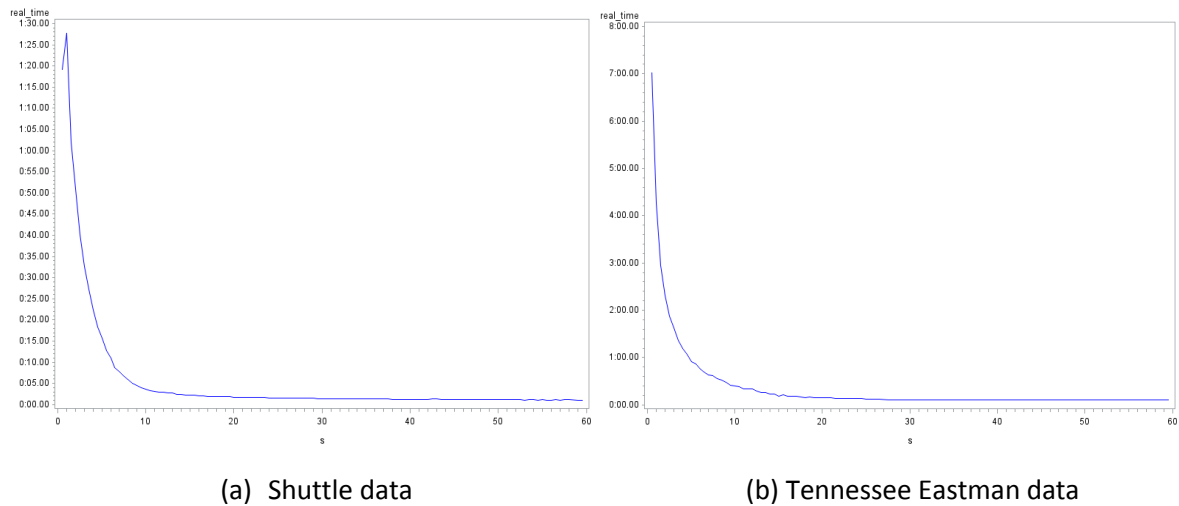

(a)  Shuttle data                         (b)  Tennessee Eastman data

Figure 1. SVDD computational time versus $s$ for two large multidimensional data sets.

| Data set | Shuttle | Tennessee Eastman |
|---|---|---|
| Number of observations | 45,586 | 108,000 |
| Number of variables | 9 | 41 |
| Total computational time, min:sec | 10:55 | 43:44 |

Table 1. Summary of full peak method performance for two large data sets.

This paper proposes an extension of the peak method for the case of large multivariate data. The method, called the sampling peak criterion method, can be used when the training data set is too large to use full SVDD training. The proposed method provides good results when applied both to small two-dimensional data sets (where the full peak method can also be used) and to large multidimensional data sets (where the full peak method might be either not practical or not feasible).

The rest of the paper is organized as follows. Section 2 presents the extension of the peak method to select optimal $s$ for large data sets. We demonstrate its performance on several small two-dimensional data sets of known geometry, and we compare it with the full peak method. Section 3 explores the method's performance on real-life large high-dimensional data. Section 4 discusses alternative methods of selecting optimal $s$ and their performance on the data sets used in the previous two sections. Section 5 concludes and provides avenues for further research.

---

[3] Detailed description of both data sets is given in Section 3.

## 2.  Sampling Peak Method

This section describes an algorithm that is used in the sampling peak criterion method (also called here the sampling peak method). We then report its performance on three small two-dimensional data sets of known geometry and compare it with the full peak method.

## 2.1.  Description of the Sampling Peak Method

The peak method described in Kakde et al. [1] relies on an SVDD training algorithm that uses the entire training data set. A large training data set can have too many observations to allow effective use of a standard SVDD training algorithm, because the time required to solve the quadratic programming problem is directly related to the number of observations in the training data set, as demonstrated by Chaudhuri et al. [2]. For large data sets, Kakde et al. [1] used a small random sample as a training sample for the peak method application. It can be argued that a single small random sample might not be sufficient to fully characterize a large multidimensional data set.

A sampling SVDD method was proposed by Chaudhuri et al. [2]. That method iteratively learns the training data set description by computing the SVDD on independent random samples that are selected with replacement. The method is shown to provide a fast and nearly identical data description as compared to using an entire training data set in one iteration. We used the sampling SVDD method to create an extension of the peak method; the extension, called the sampling peak approach, can be applied to select an optimal value of $s$ for large multidimensional data sets. The sampling peak approach for a large training data set of $N$ observations can be outlined as follows:

1. Start with a sample size $n_0 < N$ of the large data set.
2. For each sample size $n_i$ (where $i \geq 0$), run the sampling SVDD method for a range of values of $s$ from $s_{min}$ to $s_{max}$ with a step $\Delta s$. Obtain the value of the optimal objective function (OOF) for each run.
3. Compute the first difference of the OOF with respect to $s$ (as an approximation of the first derivative).
4. Smooth the first difference of the OOF by using a penalized B-spline.
5. Find the optimal value $s^i_{opt}$ as the first extremum of the smoothed first difference.
6. Increase the sample size to $n_{i+1} = n_i + \Delta n$.
7. Repeat steps 2–6 until convergence is reached for $s^i_{opt}$ . Convergence is defined as the following condition: when $\left\| s^i_{opt} - s^{i-1}_{opt} \right\| \leq \epsilon_s \left\| s^{i-1}_{opt} \right\|$ for $u$ consecutive iterations, where $\epsilon_s > 0$ and $u \in \mathbb{N}$ are pre-defined.

We first applied the sampling peak method on several two-dimensional data sets of known geometry. Each of the data sets is relatively small in order to allow full SVDD estimation; however, we used the sampling peak method to demonstrate its applicability and to compare it with the full peak method. Star-shaped data was analyzed first, followed by three clusters and banana-shaped data. For each type of data, the full peak method similar to Kakde et al. [1] is used as a starting point, followed by the sampling peak method. We used an outlier fraction of $f = 0.001$ in each case.

## 2.2. Star-Shaped Data: Full Peak Method

Star-shaped data is a two-dimensional data set of 582 observations; the scatter plot of the data is depicted in Figure 2(a). As was noted in Section 1.1, for a fixed value of the outlier fraction $f$, the number of support vectors (NSV) that are generated by SVDD training is a function of the Gaussian kernel parameter $s$. We performed full SVDD on the data for values of $s$ from 0.05 to 10 with a step of 0.05 while keeping $f$ = 0.001. The resulting NSV as a function of $s$ can be seen in Figure 2(b). For $s \leq 0.05$, NSV = 582; that is, every observation in the data set is a support vector. As $s$ increases, NSV declines until it reaches the minimum of 5 (the points on the tips of the star vertices) for $s \geq 3$.

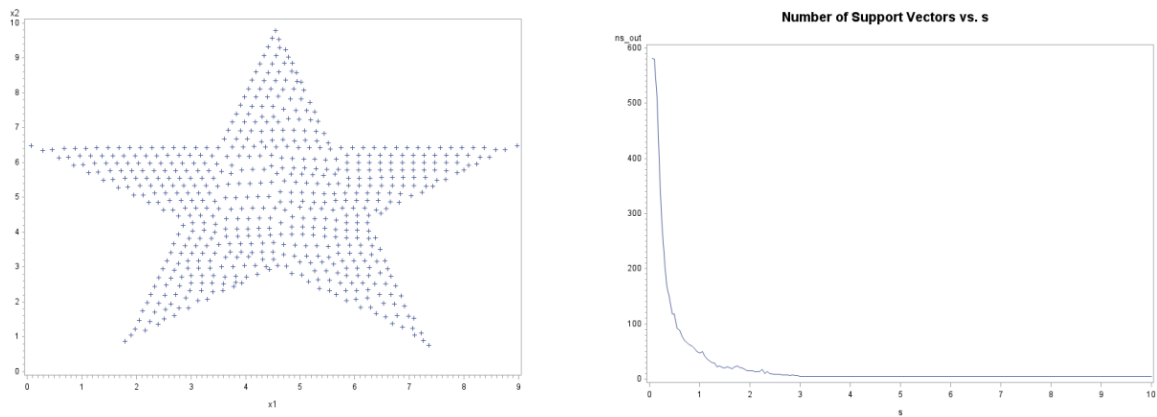

(a) Star-shaped data            (b) Number of support vectors as a function of s

Figure 2. Star-shaped data and number of support vectors as a function of s.

In order to judge how a value of $s$ affects the shape of the boundary that we observe around the data, we used the same values of $s$ as in the previous exercise and performed SVDD scoring on a 200 x 200 grid. The scoring results for four different values of $s$ are shown in Figure 3. For small values of $s$—such as $s$ = 0.3 in Figure 3(a)—the number of support vectors is comparable with the total size of the data. In addition, the boundary is jagged, with many individual data points being classified as boundary points, and in many cases, the space between any two neighboring data points (even those that are on the inside of the star shape) being classified as outside points. As $s$ increases — as in Figure 3(b) where $s$ = 0.9 — the boundary starts conforming to the shape of the data, with less space inside the data being classified as outside. However, as $s$ is increased further, the boundary starts losing its true star shape, as in Figure 3(c), until it ultimately becomes spherical for $s \geq 3$, as in Figure 3(d).

The value of an optimal objective function versus s and its first difference with respect to s are provided in Figure 4. For a particular outlier fraction $f$, the optimal objective function value is an increasing function of $s$. The first difference of the optimal objective function increases with $s$ for small values of $s$ up to around $s$ = 0.9; after that, it starts declining as $s$ increases , thus indicating that the first critical point of the first difference is around $s$ = 0.9. Kakde et al. [1] argue that the first critical point of the first derivative (approximated by the first difference) of the optimal objective function produces a value of $s$ that provides a good data boundary. Indeed, as can be seen from Figure 3, the boundary conforms to the general geometric shape of the data around $s$ = 0.9. Kakde et al. [1] use the second derivative of the optimal objective function value with respect to *s* for choosing a value of *s* where the second derivative (approximated by the second difference) reaches 0 for the first time. This criterion has been coined the peak method by Kakde et al. [1]. In the current paper, we use instead the first optimum of the first derivative of the optimal objective function value. Both methods are mathematically equivalent for a smooth twice-continuously differentiable function; however, when derivatives are approximated by the differences, using the second difference produces unsatisfactory results for higher-dimension data.

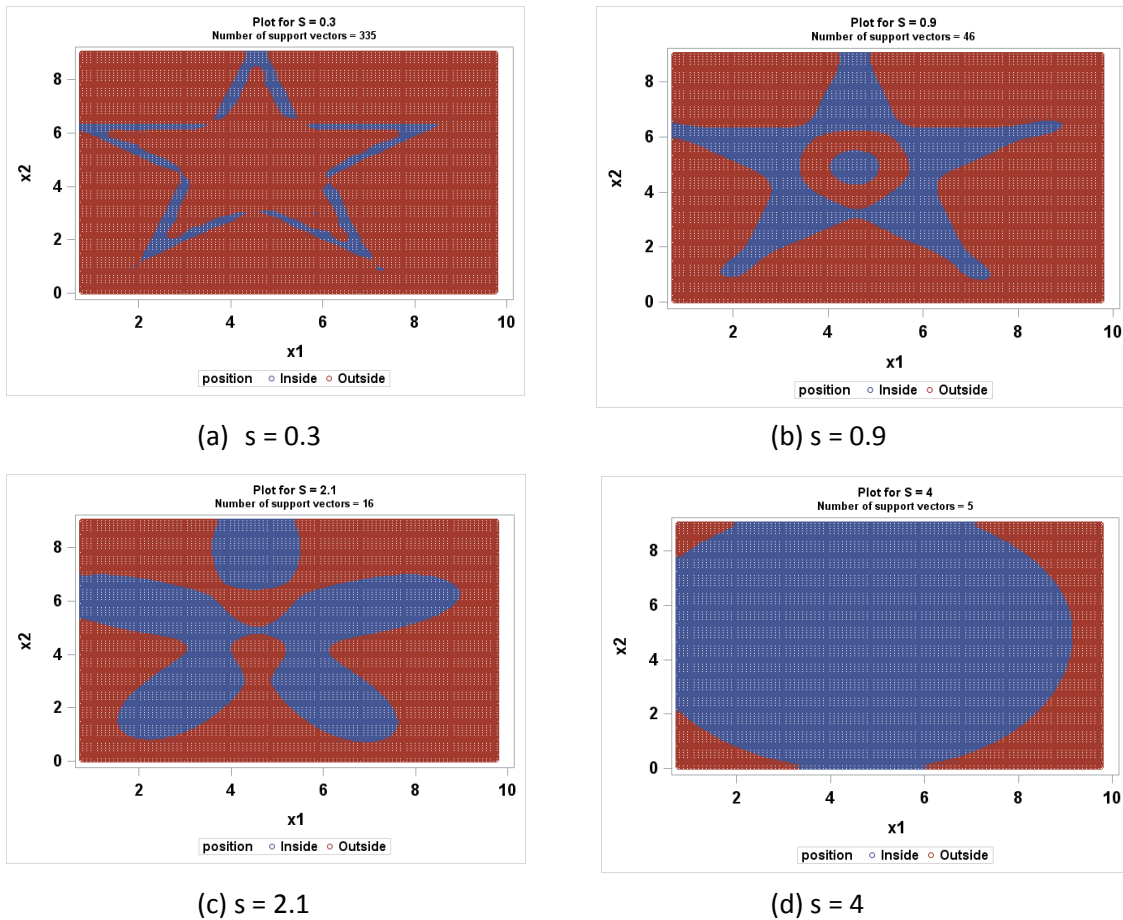

Figure 3. Scoring results for star-shaped data on a 200 x 200 grid. Blue indicates inside points, and red indicates outside points as generated by the scoring.

To find the value of the first critical point for the first derivative, we fitted a penalized B-spline by using the TRANSREG procedure from SAS/STAT® software. The fit is displayed in Figure 5. The first critical point for the smoothed value of the first difference is reached at *s* = 0.9.

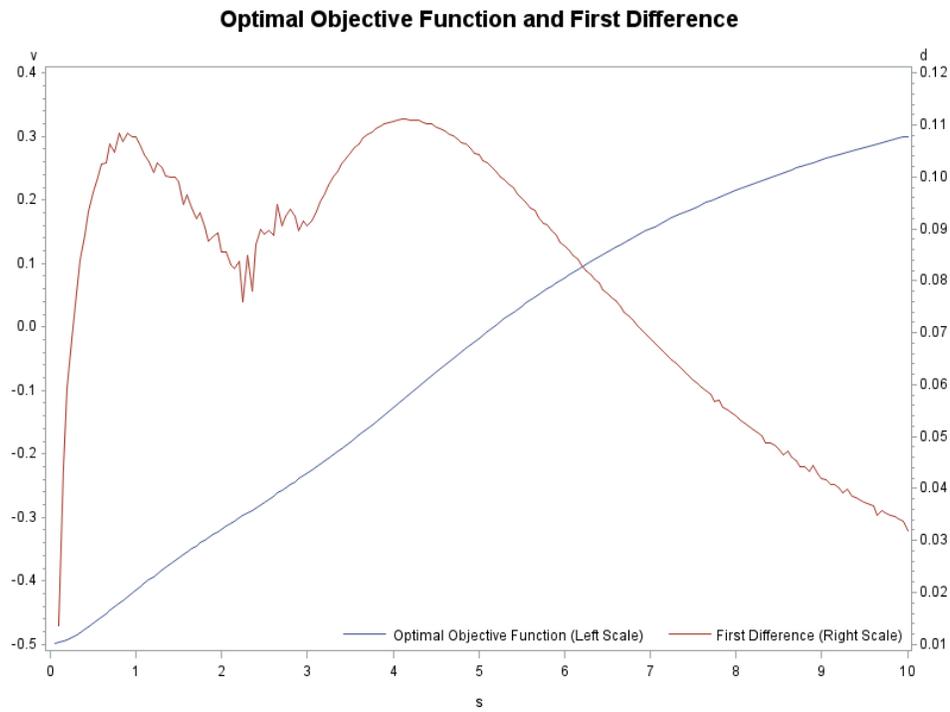

Figure 4. Optimal objective function and its first difference with respect to *s* for star-shaped data.

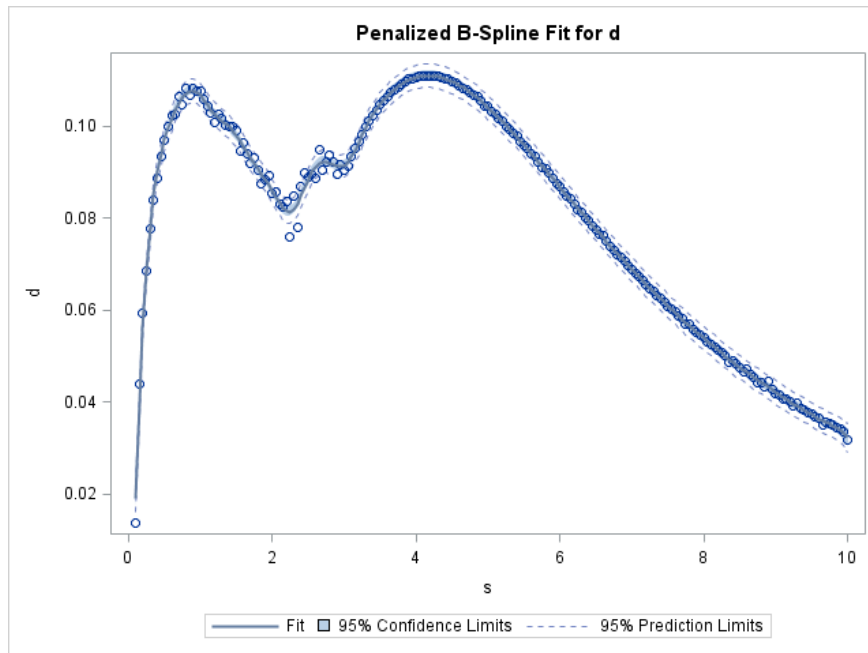

Figure 5. Penalized B-spline fit for the first difference: star-shaped data.

## 2.3.    Star-Shaped Data: Sampling Peak Method

Sampling SVDD on star-shaped data was performed according to the methodology offered by Chaudhuri et al. [2]. Figure 6(a–c) shows results for the first difference of the optimal objective function value with respect to s for three different sampling sizes: 29, 116, and 233 observations (5%, 20%, and 40% of the data, respectively). In addition, Figure 6(d) superimposes these three graphs and the results for full SVDD training. For each sample size, we ran sampling SVDD training for values of $s$ from 0.05 to 10 with a step of 0.05. As you can see in the Figure 6, although the overall shape of the first difference has larger oscillations from a smooth curve for smaller sample sizes, it is still similar for different sample sizes. We used a penalized B-spline to find the value of $s$ for the first extremum of the smoothed first difference; the results are shown in Table 2.

| Sample size | Sample size % | $s_{opt}$ |
|---|---|---|
| 29 | 5% | 1.1 |
| 116 | 20% | 0.9 |
| 233 | 40% | 0.9 |

Table 2. Sampling peak method results for different sample sizes, star-shaped data.

As the next step, we performed sampling SVDD training for star-shaped data for sample sizes from 29 (5% of the data) to 582 (100%) with a step of 6 (1%). For each sample size, we ran sampling SVDD training for values of $s$ from 0.05 to 10 with a step of 0.05. Then, for each sample size, we obtained the first difference of the optimal objective function, smoothed it by using a penalized B-spline, and then found $s_{opt}$ as an argument of the first optimum of the smoothed first difference. The value of $s_{opt}$ for each sample size is displayed in Figure 7. As you can see from the Figure 7, starting from sample size of 99 (17% of the data), the value of $s_{opt}$ converges to a range [0.85, 0.9], indicating that the overall optimal $s$ for the sampling peak method is between these two values.

## 2.4.    Three-Cluster Data

Next, we used our technique on three-cluster data. The data consist of 276 two-dimensional observations arranged into three distinct clusters as shown in Figure 8(a). Figures 8(b–c) show the results of scoring on a 200 x 200 grid after full SVDD training for different values of $s$. Similar to what was observed in the case of star-shaped data, we found that for small values of $s$—such as in Figure 8(b) where $s$ = 0.5—the data boundary is too jagged. As $s$ increases, the data boundary becomes smoother; around the value of $s$ = 2.1, it resembles the shape of the data with three distinct clusters (Figure 8(c)). With further increase in $s$, scoring clusters merge into one (Figure 8(d)).

Figure 9(a) provides the results for the optimal objective function and its first difference for full SVDD training for the range of $s$ values from 0.05 to 10 with a step of 0.05. The graph for the first difference illustrates that the first critical point for the first difference is around $s$ = 2.1, which is confirmed by using a penalized B-spline. Figure 9(b) shows the value of the first difference with respect to $s$ for sampling SVDD training for the sample sizes of 14, 55, 110, and 276 observations (5%, 20%, 40%, and 100% of the data, respectively).

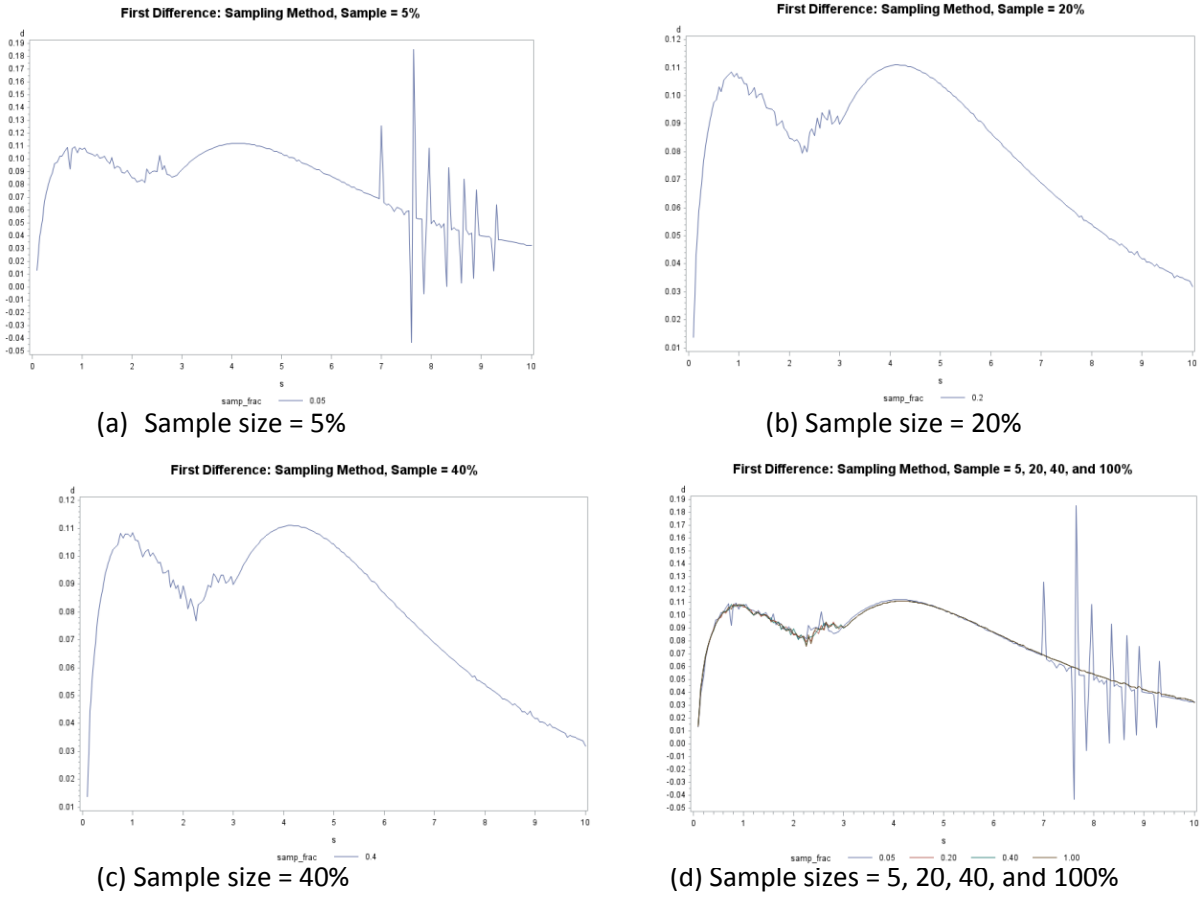

Figure 6. First difference of the optimal objective function for sampling SVDD: star-shaped data.

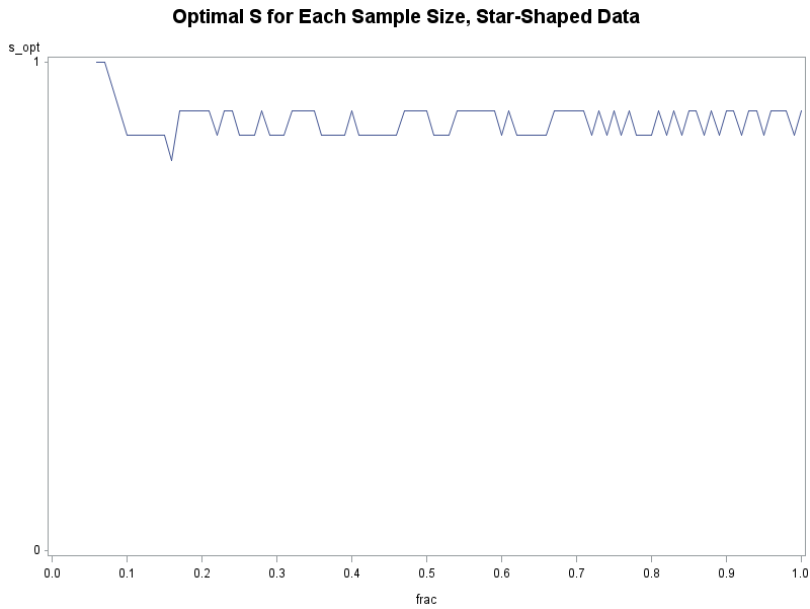

Figure 7. Optimal *s* for each sample size for star-shaped data.

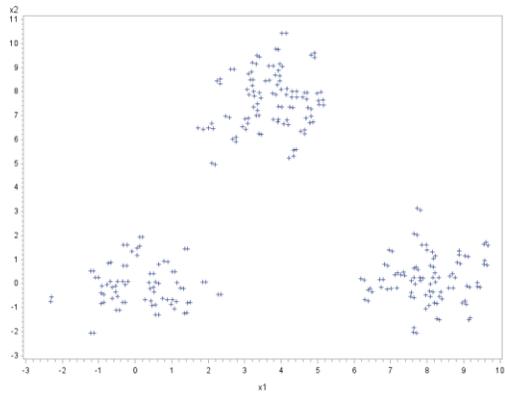

(a) Scatter plot of three-cluster data

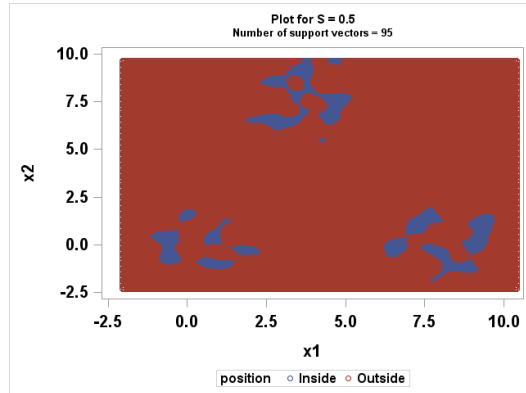

(b) s = 0.5

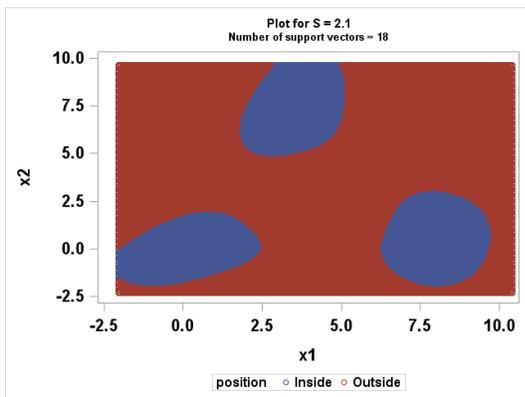

(c) s = 2.1

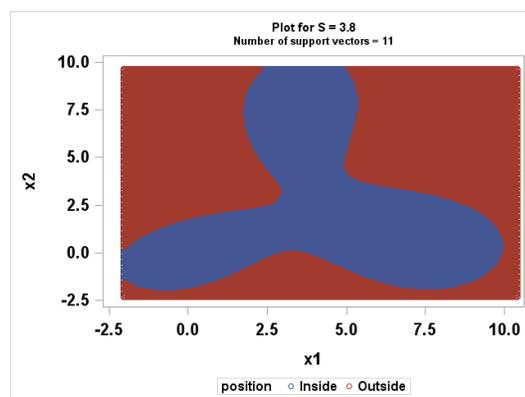

(d) s = 3.8

Figure 8. Scoring results for three-cluster data on a 200 x 200 grid for different values of *s*. Blue indicates inside points, and red indicates outside points as generated by the scoring.

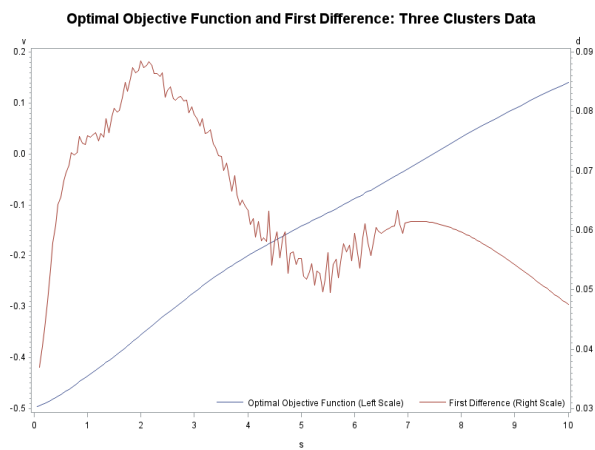

(a) Optimal objective function and its first difference for full SVDD

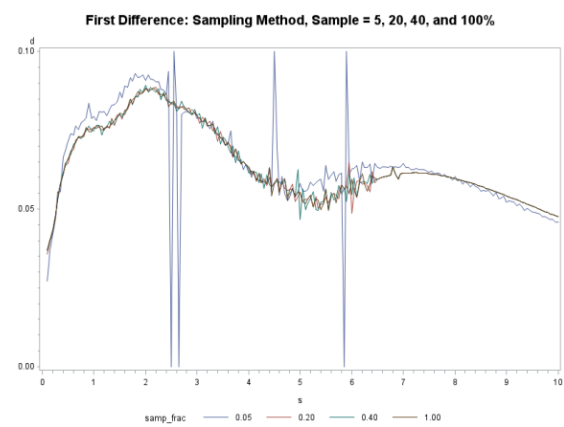

(b) First difference of the optimal objective function for sample sizes of 5, 20, 40, and 100%

Figure 9. First difference for full SVDD and sampling SVDD for three-cluster data.

We ran sampling SVDD with sample sizes from 14 to 276 with a step of 3 (5%, 100%, and 1% of the data, respectively). For each sample size, the value of $s$ was changed from 0.05 to 10 with a step of 0.05. The first difference of the optimal objective function for each sample size was smoothed using a penalized B-spline, and the $s_{opt}$ value as an argument of the first optimum of the smoothed first difference was stored. Figure 10 shows the results of $s_{opt}$ versus sample size; you can see that $s_{opt}$ converges to 2.1 starting from the sample size of 47 (17% of the data).

## 2.5.  Banana-Shaped Data

Banana-shaped data is a two-dimensional data set of 267 observations. The scatter plot of the data is shown in Figure 11. The data were trained for a range of values of $s$ from 0.05 to 10 with a step of 0.05. The scoring results for select values of $s$ on a 200 x 200 grid are shown in Figure 12. The value of the optimal objective function and its first difference versus s for full SVDD is depicted in Figure 13(a). The value of the first difference of the optimal objective function for sampling SVDD for select sample sizes (13, 53, 107, and 267; that is, 5%, 20%, 40%, and 100% of the data, respectively) is shown in Figure 13(b).

Next, the same training exercise was performed using sampling SVDD for sample sizes from 13 to 267 with a step of 3 (5%, 100%, and 1% of the data, respectively). When you use a penalized B-spline to smooth the data, you must specify the number of internal knots — that is, the number of equidistant values of the $x$ variable where spline pieces meet. By default, the TRANSREG procedure assumes 100 knots for penalized B-spline regression. Figure 14(a) shows the results of using the default value of 100 knots for a penalized B-spline on full banana-shaped data. The first maximum for the smoothed first difference of the optimal objective function is reached at $s_1^{max}$ = 0.4. However, you can see that soon afterward, the smoothed function reaches a local minimum at $s_1^{min}$ = 0.65 and soon after that, reaches the second local maximum at $s_2^{max}$ = 0.8. The values of the smoothed function at all three extrema are relatively close to each other: $f\left(s_1^{min}\right)/f\left(s_1^{max}\right) = 0.986$ and $f\left(s_1^{min}\right)/f\left(s_2^{max}\right) = 0.997$. Therefore, you could argue that the default number of knots leads to overfitting and the smoothing should result in one local maximum in this $s$ range. By reducing number of knots to 40, the penalized B-spline generates one local maximum at $s$ = 0.65 as demonstrated in Figure 14(b). This result is supported by visual observation as in Figure 12: the scoring result on a 200 x 200 grid produces a better boundary and conforms to the shape of the data better for s = 0.65 (Figure 12(c)) than it does for s = 0.4 (Figure 12(b)).

The optimal $s$ that is obtained by using sampling peak method for each sample size is plotted in Figure 15. Here a solid line is for the optimal $s$ obtained by running a penalized B-spline with the default value of 100 knots, and the dashed line is the same exercise with 40 knots. In both cases, you see convergence starting from a sample size of 36%: in the former case, the optimal $s$ converges to a range of [0.4, 0.45], and in the latter case, it converges to a range of [0.65, 0.7].

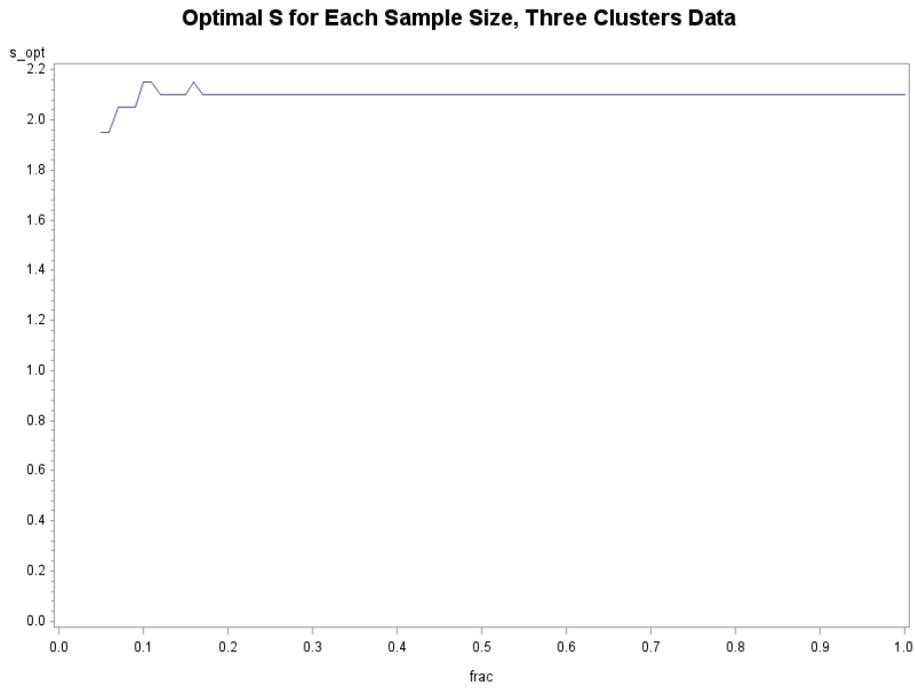

Figure 10. Optimal value of *s* for different sample sizes for three-cluster data.

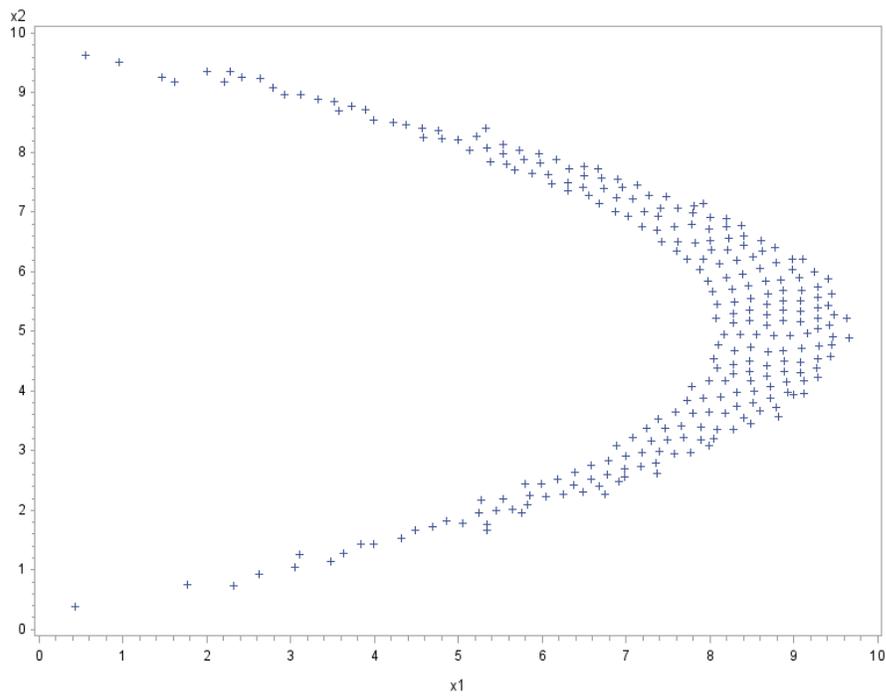

Figure 11. Scatter plot of banana-shaped data.

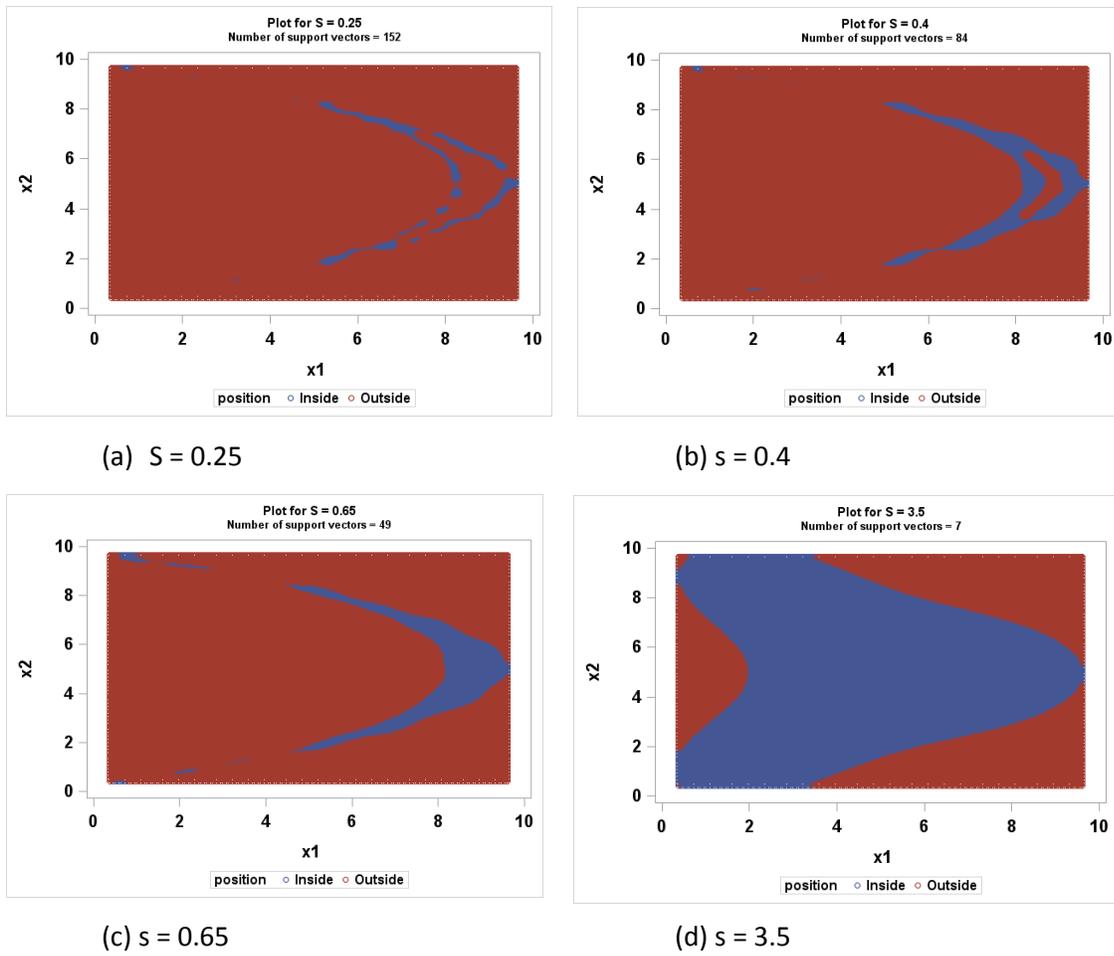

(a)  S = 0.25

(b) s = 0.4

(c) s = 0.65

(d) s = 3.5

Figure 12. Scoring results on a 200 x 200 grid for selected values of s for banana-shaped data.

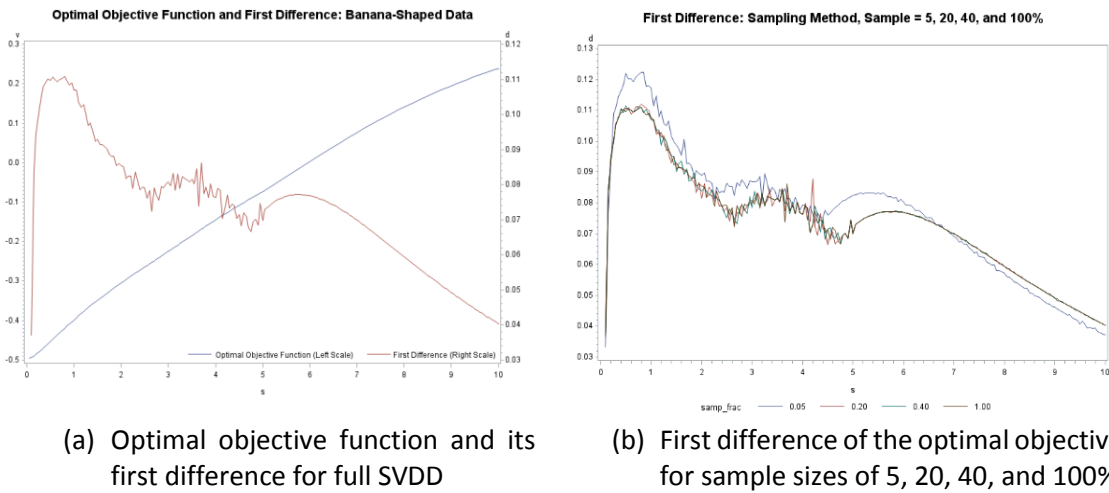

(a)  Optimal objective function and its first difference for full SVDD

(b)  First difference of the optimal objective function for sample sizes of 5, 20, 40, and 100%

Figure 13. First difference for full SVDD and sampling SVDD for banana-shaped data.

# 3.   Analysis of High-Dimensional Data

Two-dimensional data can provide visual guidance for judging the value of $s$ that is selected by the peak criterion. For higher-dimensional data, however, such guidance is not possible, and alternative methods must be used to evaluate the performance of the peak criterion in selecting the proper value of $s$. This paper uses the $F_1$ measure as an alternative criterion for evaluating the peak method. The $F_1$ measure is defined as follows:

$$F_1 = \frac{2 \times Precision \times Recall}{Precision + Recall},$$

where

$$Precision = \frac{True\ Positives}{True\ Positives + False\ Positives}$$

$$Recall = \frac{True\ Positives}{True\ Positives + False\ Negatives}$$

The following two sections apply sampling peak method to two large multivariate data sets: Shuttle and Tennessee Eastman.

## 3.1.   Shuttle Data Set

The Shuttle data set [3] consists of nine numeric attribute variables and one class attribute variable. It has 58,000 observations, 80% of which belong to class one. For our analysis, we created a training data set by selecting a random sample of 2,000 observations that belong to class one. The rest of the data (56,000 observations) was used for scoring to determine whether the model could successfully classify whether an observation belongs to class one. Training and scoring were performed for the range of values of $s$ from 1 to 100 with a step of 1. The model performance was evaluated using an $F_1$ measure for each $s$. The graph of $F_1$ measured against $s$ is shown in Figure 16. $F_1$ reaches its maximum for $s = 17$.

To apply the sampling peak method on the Shuttle data, we conducted two experiments. First, sampling SVDD training was performed for a sample sizes that ranged from 100 to 800 (5% to 40% of the 2,000-observation training sample) with a step of 20 (1%). For each sample size, we varied $s$ from 0.05 to 60 with a step of 0.05. Figure 17(a) shows the value of the optimal objective function and its first difference for a sample size of 20%. Figure 17(b) depicts the penalized B-spline of the first difference of the optimal objective function for the same sample. The graph shows that the smoothed first difference has only one maximum at $s = 15.4$.

Figure 18 shows the results of the sampling peak method for the Shuttle data 2,000-observation training sample. As you can see from the graph, $s_{opt}$ converges to a range [15.3, 15.75] for a sample size of 160 (8%) and higher. For this range of values, the ratio of $F_1(s_{opt})/ F_1{}^{max}$ is 0.996 or higher. Here $F_1{}^{max} = F_1(s = 17)$, the maximum value. The fact that the sampling peak method provides an $F_1$ measure for the optimal s that is very close to the maximum, provides empirical evidence that the sampling peak method can be successfully applied to higher-dimensional data.

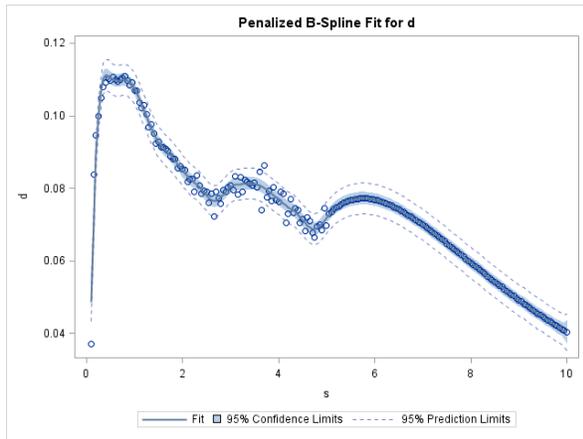 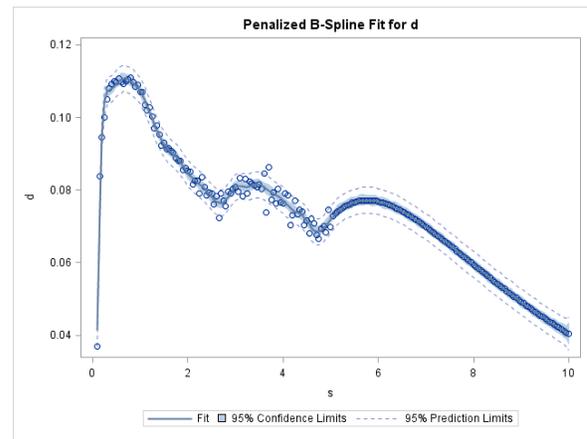

(a)  Number of knots = 100 ($s_{opt}$ = 0.4)    (b)  Number of knots = 40 ($s_{opt}$ = 0.65)

Figure 14. Penalized B-spline for the first difference for different number of knots; banana-shaped data.

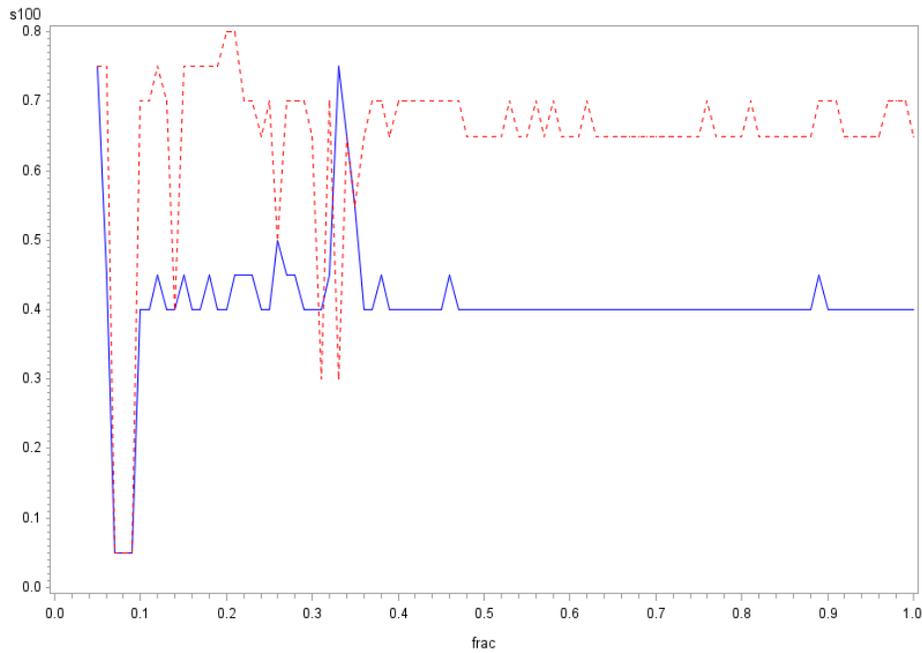

Figure 15. Optimal value of *s* for different sample sizes for banana-shaped data. Solid line: 100 knots; dashed line: 40 knots for penalized B-spline smoothing.

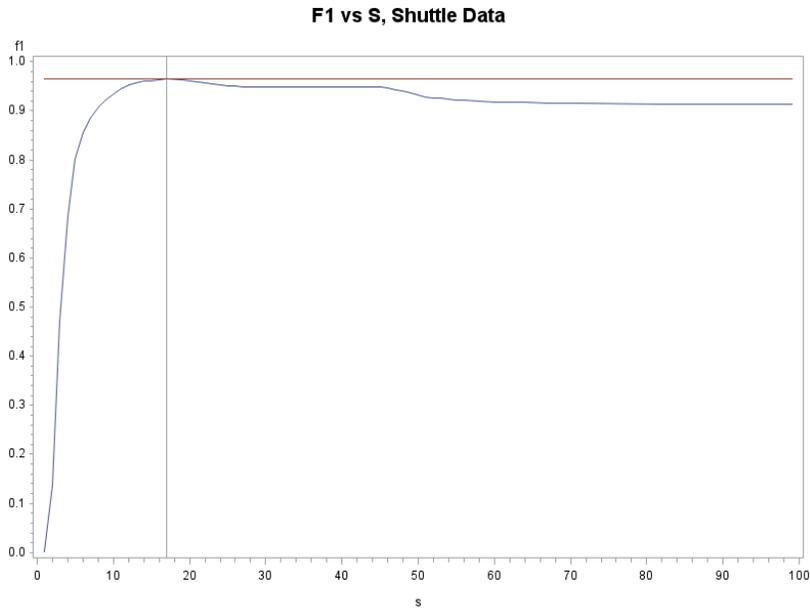

Figure 16. $F_1$ measure versus s for the Shuttle data.

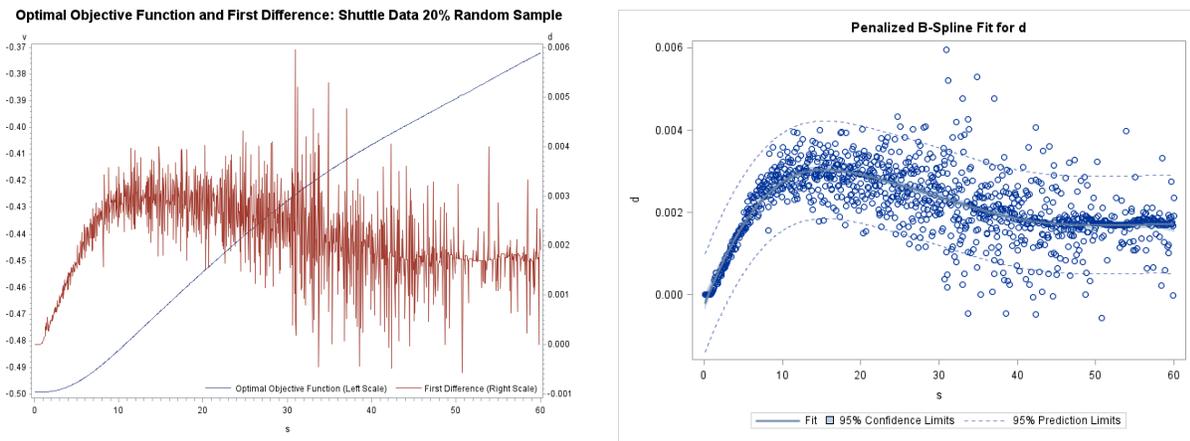

Figure 17. First difference of the optimal objective function and its fit for penalized B-spline for 20% sample of the Shuttle data.

As the next step, we applied the sampling peak method to the entire Shuttle data of class one, which consists of 45,586 observations. We performed the sampling SVDD training with a sample size from 46 to 3,647 observations (0.1% to 8% of the entire class-one data) with a step of 46 and $s$ varying from 0.05 to 60 with a step of 0.05 for each sample size. The optimal $s$ for each sample size for this exercise is shown in Figure 19. The optimal $s$ converges to a range [8.75, 9] starting from a sample size of 2,462 observations (5.4% of the data). For this range of $s$ values, the ratio of $F_1(s_{opt})/ F_1^{max}$ is 0.956.

The difference between the optimal values of $s$ that are suggested by the two experiments can be attributed to the fact that a small random sample might not be sufficient to capture the properties of a large data set.

## 3.2.   Tennessee Eastman Data

In this section, we experimented with Tennessee Eastman data. The data set is generated by Matlab code [4], and it simulates an industrial chemical process. The model can generate results for both normal operations and for 20 types of faults. The data set contains 41 variables, 22 of which are measured continuously (on average every 6 seconds), and the other 19 of which are sampled at a specified interval (either 0.1 or 0.25 hours). For our exercise, the data were interpolated to have 20 observations per second. The resulting training data set contains 108,000 observations for normal operations. To generate the results for the $F_1$ measure, the random sample of 2,000 observations from the normal-operations data set was selected for training. For scoring, another random sample of 2,000 observations was selected from combined data for normal operations and for fault 1. The SVDD scoring was performed to determine whether the model could successfully classify if an observation belongs to normal operations. We performed training and scoring for a range of $s$ values from 1 to 100 with a step of 1.

The results for the $F_1$ measure are shown in Figure 20. The absolute maximum is reached at $s$ = 30. It is also worth noting that another local maximum is reached earlier at $s$ = 22, with the $F_1$ value being very close to the absolute maximum: $F_1(22)/F_1(30) = 0.998$.

For the sampling peak exercise, we ran sampling SVDD training for sample sizes from 108 to 4,320 observations (0.1% to 4% of 108,000-observation normal-operation data set) with a step of 108. For each sample size, the value of $s$ was changed from $s_{min}$ = 0.05 to $s_{max}$ = 200 with a step of 0.05. To determine whether the results were influenced by the choice of the maximum value of $s$, we fitted a penalized B-spline that had three different values of $s_{max}$: 60, 100, and 200. For a proper comparison, we chose the number of knots so that the knots are located at the same values of $s$ for each selection of $s_{max}$. The number of knots were 100, 167, and 333, respectively. The results for optimal value of $s$ against the sample size for three values of $s_{max}$ are shown in Figure 21. It can be inferred from the graph that the choice of $s_{max}$ does not influence convergence: in all three cases, $s_{opt}$ converges to a range [21.1, 21.95] starting from a sample size of 540 (0.5% of the training data). The range of optimal values of $s$ that is suggested by the sampling peak method is at least 99.5% of the maximum $F_1$ value: $F_1(21)/F_1(30) = 0.995$ and $F_1(22)/F_1(30) = 0.998$. These results provide empirical evidence of the applicability of the sampling peak method for large high-dimensional data.

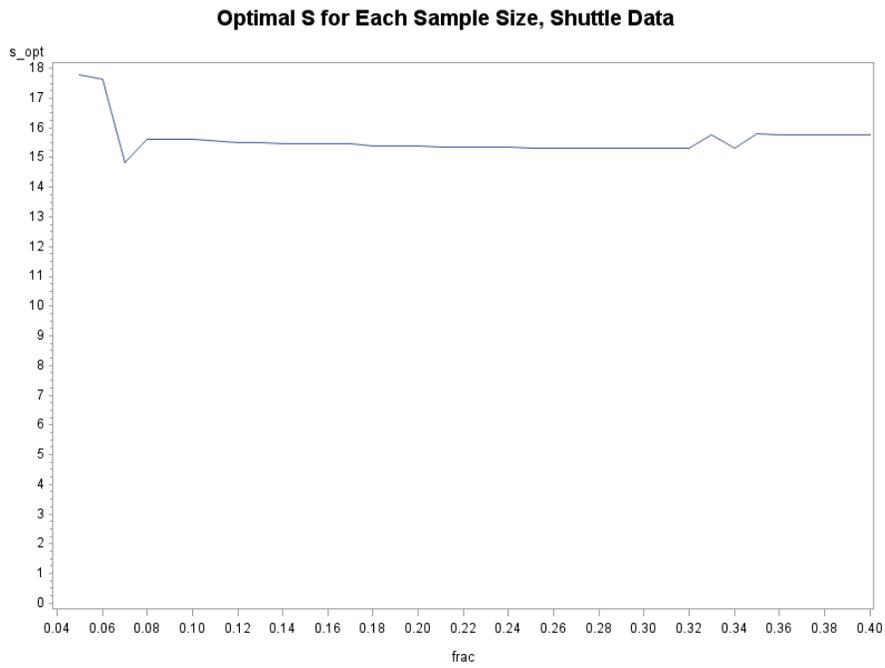

Figure 18. Optimal *s* against sample size for a Shuttle data training sample of 2000 observations.

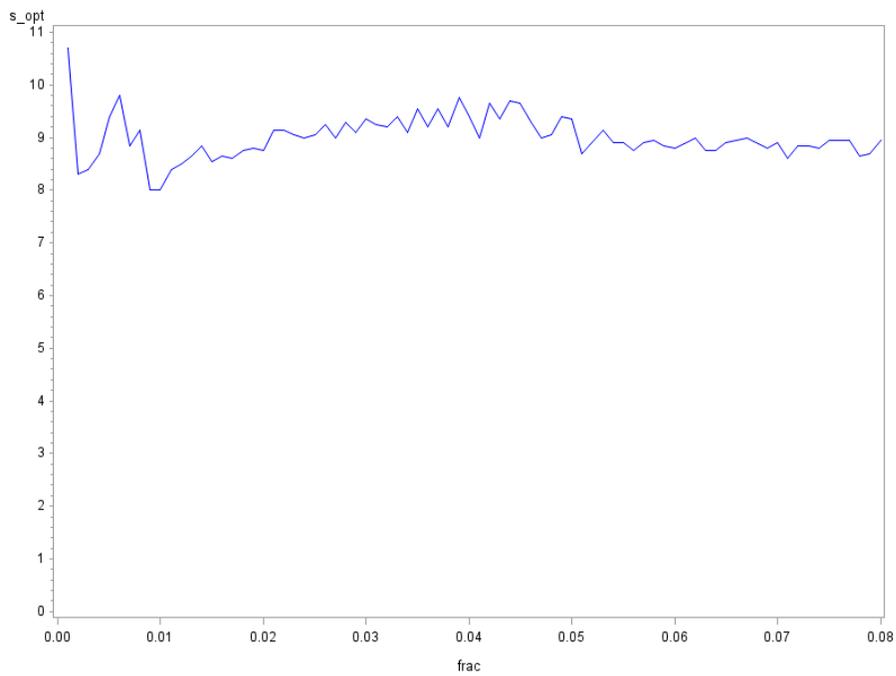

Figure 19. Optimal *s* against sample size for a sample from the entire Shuttle data of class one.

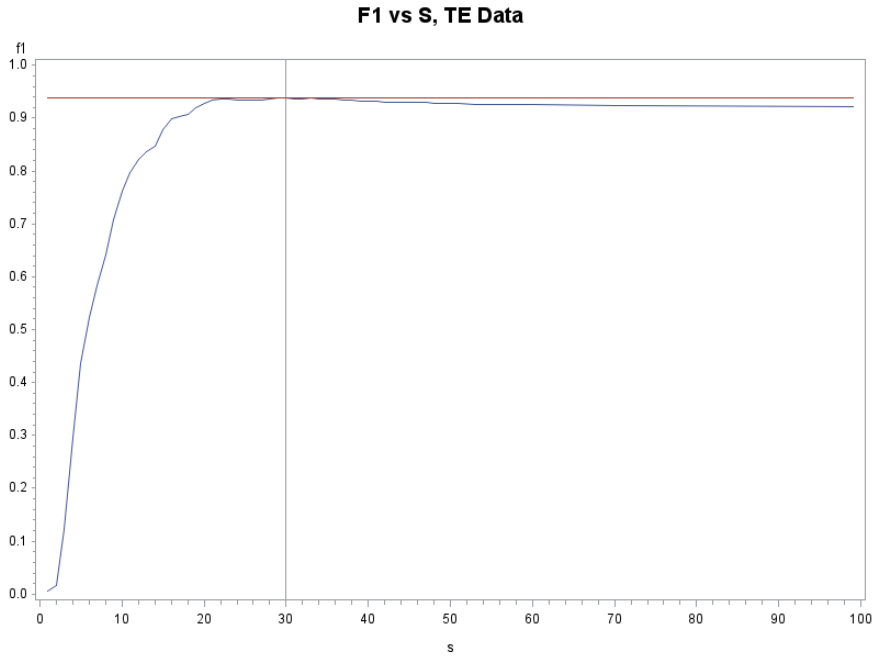

Figure 20. Optimal *s* based on *F₁* criterion: s = 30 (Tennessee Eastman data).

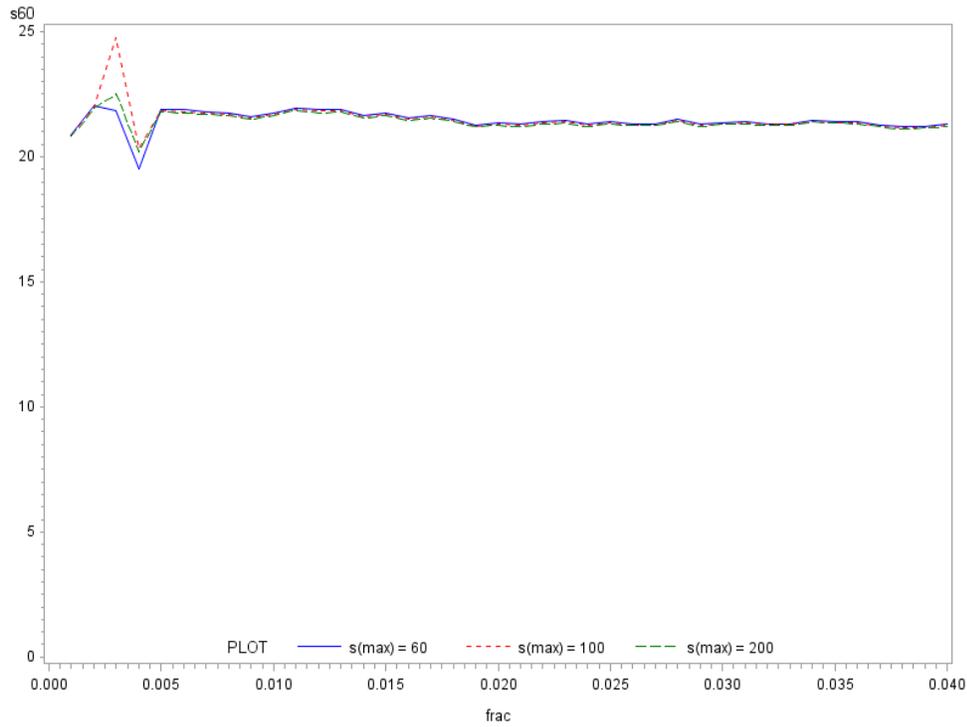

Figure 21. Optimal *s* against sample size having different values of *sₘₐₓ* for Tennessee Eastman data.

## 3.3. Performance

An important feature of the sampling peak method is its speed. To demonstrate its advantage, we timed the sampling peak method for samples of different sizes for the Shuttle and TE data. The parameters of the runs are summarized in Table 3. For each sample size, we recorded the total time to run sampling SVDD for all $s$ from $s_{min}$ to $s_{max}$ (120 runs for each sample size). The results are displayed in Figure 22 for Shuttle data, and Figure 23 for TE data. The horizontal line on each graph is the time to run the full peak method for each data set. As you can infer from the graph, the sampling peak method outperforms the full peak method up to relatively large sample sizes and well past its point of convergence. The sampling peak method is faster for the Shuttle data for sample sizes smaller than 17,186 observations (37.7% of the full training data set), and for the TE data for sample sizes smaller than 37,908 observations (35.1% of the full training data set). At the same time, as was shown in Sections 3.1 and 3.2, the sampling peak method converges to a narrow range that contains $s_{opt}$ starting from much smaller sample sizes: 2,462 observations (5.4%) for the Shuttle data and 540 observations (0.5%) for the TE data.

| Data set | $n_{min}$ | $n_{max}$ | $\Delta n$ | $s_{min}$ | $s_{max}$ | $\Delta s$ |
|---|---|---|---|---|---|---|
| Shuttle | 46 (0.1%) | 22,793 (50%) | 46 (0.1%) | 0.5 | 60 | 0.5 |
| TE | 108 (0.1%) | 43,200 (40%) | 540 (0.5%) | 0.5 | 60 | 0.5 |

Table 3. Parameters of performance runs of the sampling peak method for Shuttle and TE data. Numbers in parentheses are percentages of corresponding full data set.

## 4. Alternative Methods

In this section, we compare the performance of the sampling peak method against alternative methods of selecting the Gaussian bandwidth parameter that are published in the literature. In particular, we consider the method of coefficient of variation (CV) [5] and the method of distance to the furthest neighbor (DFN) [6]. In order to do a better comparison with the sampling peak method (including its convergence property with increasing sample size), we modified these methods to include randomization with increasing sample size. In particular, we devised the following algorithm:

1. For each sample size $n_i$ starting from $n_{min}$, draw a random sample.
2. For this random sample, find $s_{opt}$ according to the method being considered (CV or DFN).
3. Repeat steps 1–2 $M$ times; find the average $s_{opt}$ over $M$ random draws of size $n_i$.
4. Increase the sample size by an increment $\Delta n$. Repeat steps 1–3 until the selected sample size $n_{max}$ is reached.

Here $M$, $n_{min}$, $n_{max}$, and $\Delta n$ are parameters that are selected. As we increase sample size, we expect the average $s_{opt}$ to converge to a certain value, with its variance decreasing.

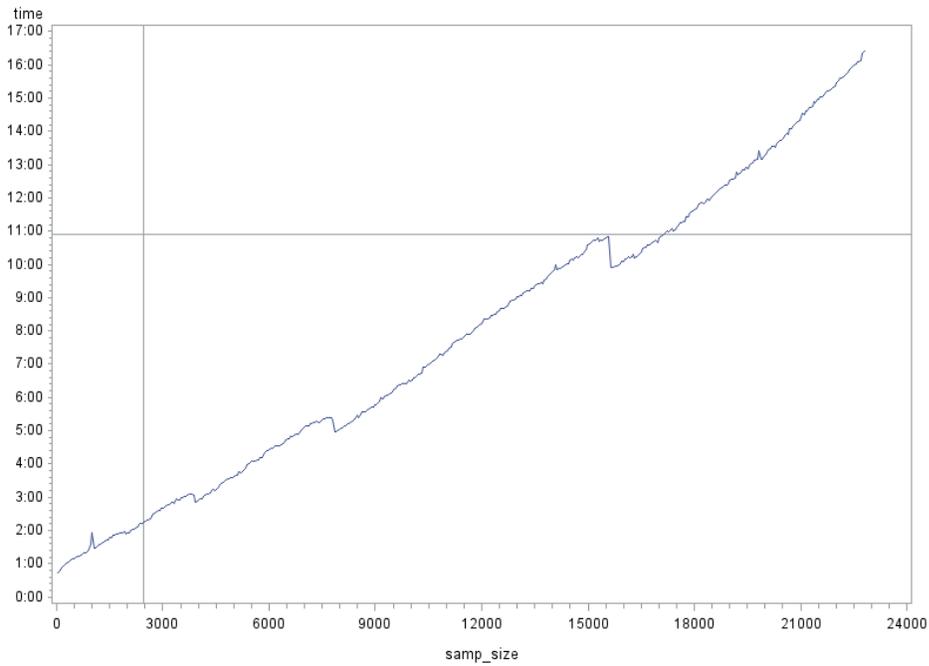

Figure 22. Time to run the sampling peak method versus sample size for Shuttle data. Horizontal line shows time to run full peak method. Vertical line shows minimum sample size for the sampling peak convergence.

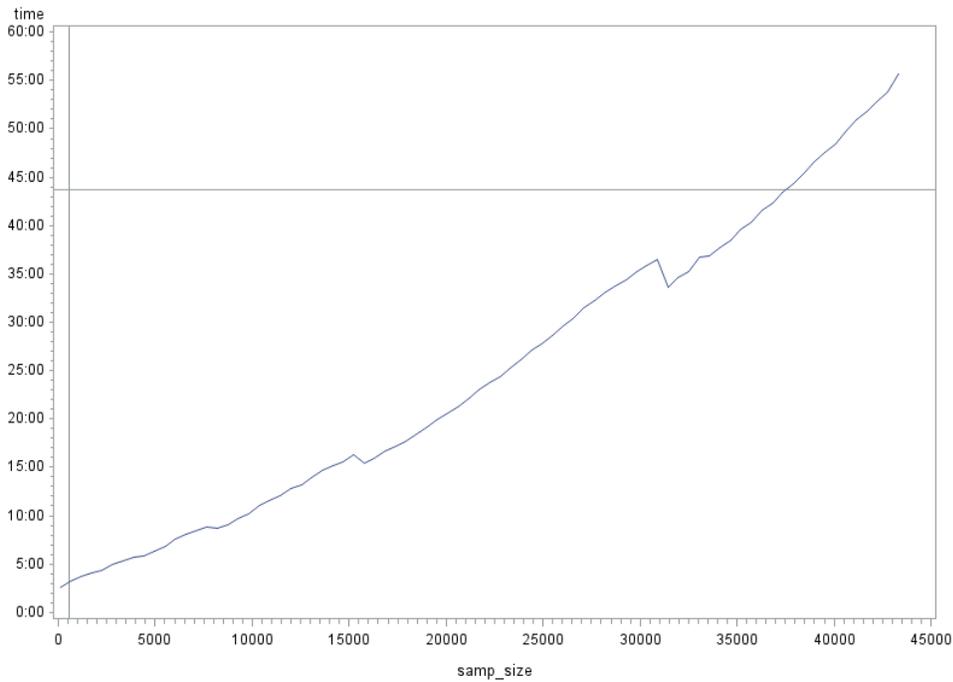

Figure 23. Time to run the sampling peak method versus sample size for TE data. Horizontal line shows time to run full Peak method. Vertical line shows minimum sample size for the sampling peak convergence.

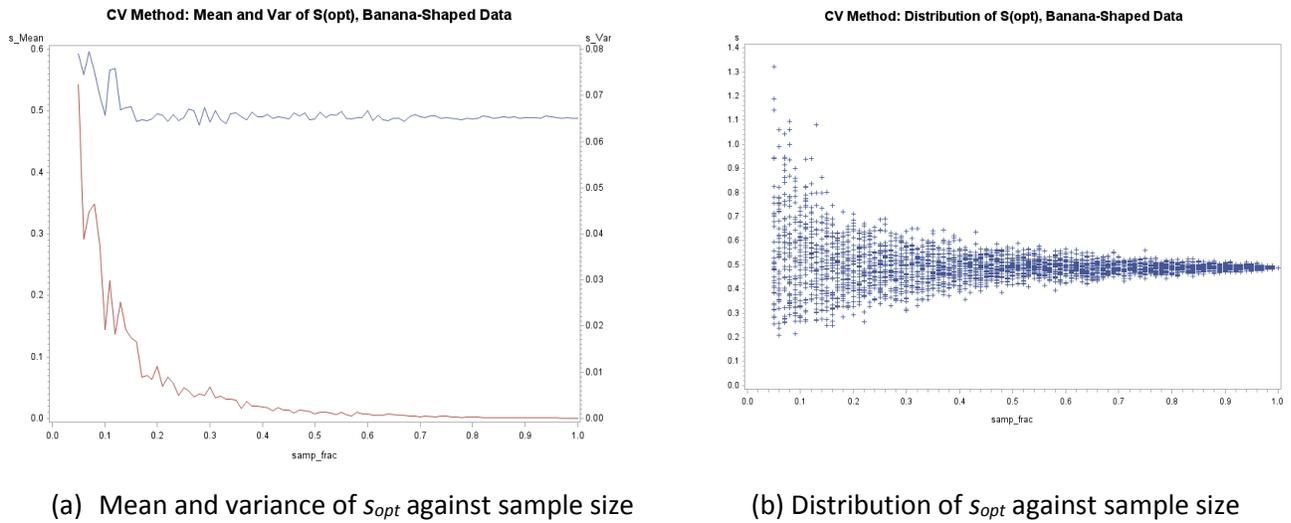

(a) Mean and variance of $s_{opt}$ against sample size

(b) Distribution of $s_{opt}$ against sample size

Figure 24. Sampling CV method for banana-shaped data.

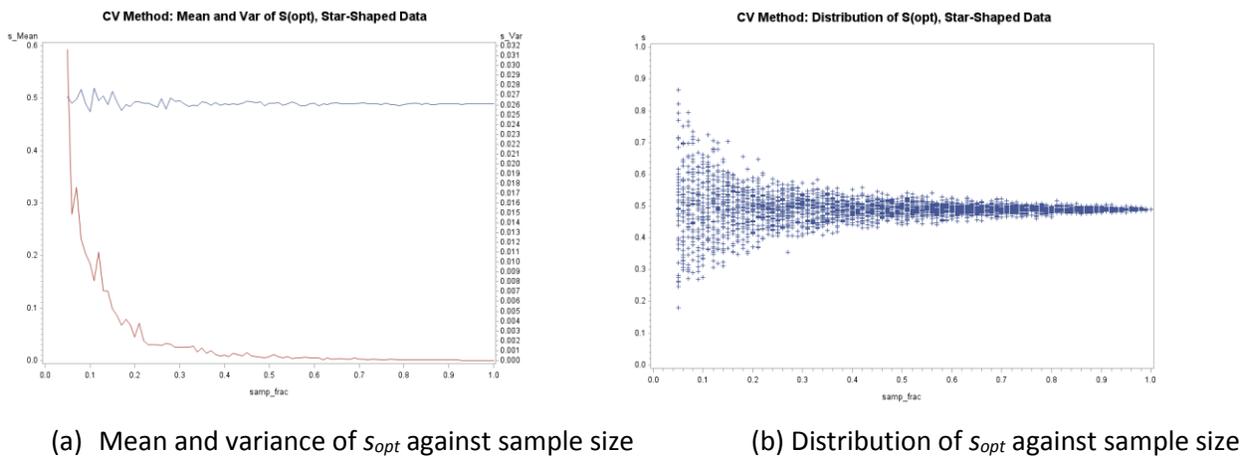

(a) Mean and variance of $s_{opt}$ against sample size

(b) Distribution of $s_{opt}$ against sample size

Figure 25. Sampling CV method for star data.

## 4.1. Method of Coefficient of Variation

The method of coefficient of variation (CV) [5] selects a value of $s$ that maximizes the coefficient of variation of the kernel matrix:

$$CV = \frac{Var}{Mean + \epsilon},$$

where *Var* and *Mean* are the variance and mean of nondiagonal elements of the kernel matrix, and $\epsilon$ is a small value to protect against division by 0 (set to 1e–6).

For small two-dimensional data sets (banana-shaped, star, and three clusters), we drew M = 40 random samples for each sample size from $n_{min}$ = 5% to $n_{max}$ = 100% of the data with a step of Δn = 1%. The results are shown in Figures 24 (for banana-shaped data), Figure 25 (for star data), and Figure 26 (for three-clusters data). Figures 24(a), 25(a), and 26(a) show the mean and the variance of optimal $s$ (against the sample size) that are proposed by sampling the CV method for the corresponding data set, whereas Figures 24(b), 25(b), and 26(b) show the actual distribution of $s_{opt}$ against the sample size. For banana-shaped and star data, the method shows good convergence as sample size increases, towards $s_{opt}$ = 0.49 for both data sets. However, for three-cluster data, the method fails to converge to a single value; variance does not decrease with increased sample size. Figure 26(b) shows two distinct values of convergence for $s_{opt}$: 0.18 and 2.28.

For the Shuttle data, the sample size was changed from 1% to 30% with a step of 1% with 40 random samples drawn for each sample size. The results are shown in Figure 27. You can infer from the actual distribution for $s_{opt}$ in Figure 27(b) that there are two distinct values of convergence.

On the Tennessee Eastman data, the sampling CV method was performed for sample sizes from 0.1% to 3.6% with a 0.1% step and 40 random sample draws for each sample size. The results depicted in Figure 28 suggest convergence to a single value for $s_{opt}$. This value is between 5 and 6, which is much smaller than the value suggested by the sampling peak method or the $F_1$ measure.

## 4.2.    Method of Distance to the Farthest Neighbor

The method of distance to the farthest neighbor (DFN) [6] finds a value of $s$ that maximizes the following objective function, which contains distances from the training data points to their farthest and their nearest neighbors,

$$f_0(s) = \frac{2}{n} \sum_{i=1}^{n} max_{j \neq i} k(x_i, x_j) - \frac{2}{n} \sum_{i=1}^{n} min_j k(x_i, x_j)$$

where n is the number of observations in training data and $k(x_i, x_j)$ is kernel distance between observations $i$ and $j$.

As in the previous exercise for small two-dimensional data, a sampling DFN was computed using 40 random samples drawn for each sample size from 5% to 100% with a step of 1%. The results are shown in Figures 29, 30, and 31 for banana-shaped, star, and three-cluster data, respectively. The method demonstrates unbiased convergence for banana-shaped and star data to the values of 2.4 and 1.7, respectively. For three-cluster data, however, the convergence behavior is biased, where the average $s_{opt}$ shows a steady decrease with increasing sample size.

For the large multivariate data sets (Shuttle and TE), the sample size was changed from 1% to 30% with a step of 1% for each data set, with 40 random samples drawn for each sample size. For the Shuttle data, the method fails to converge, demonstrating a very large variance of $s_{opt}$ for any sample size from 1 to 30% as shown in Figure 32.

For the Tennessee Eastman data, biased convergence is observed, as shown in Figure 33. The average $s_{opt}$ steadily decreases with increased sample size. The values of $s_{opt}$ are much larger than suggested by either the sampling peak method or the $F_1$ measure, with a range of values between 620 and 820.

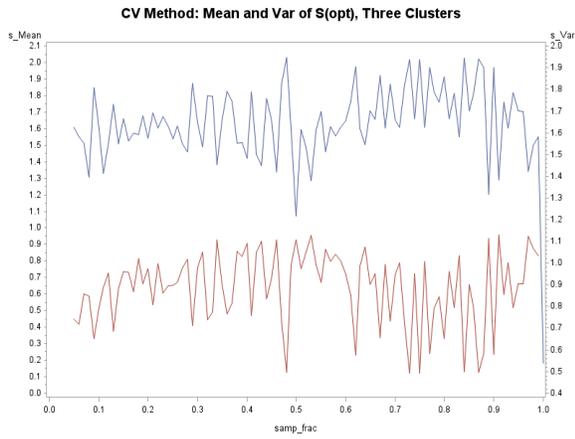 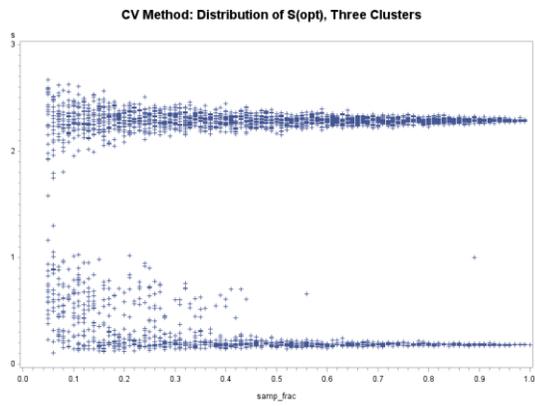

(a)  Mean and variance of $s_{opt}$ against sample size          (b)  Distribution of $s_{opt}$ against sample size

Figure 26. Sampling CV method for three-cluster data.

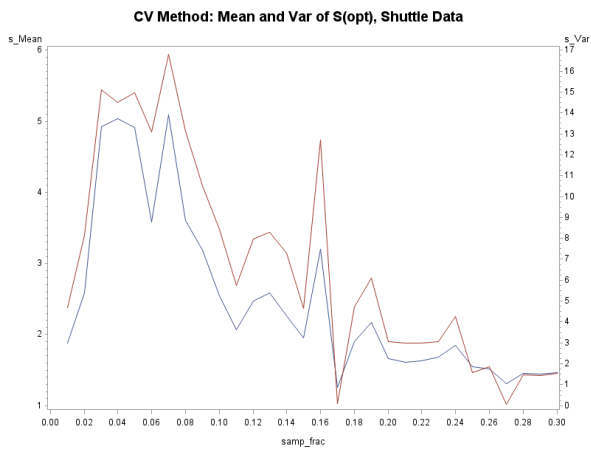 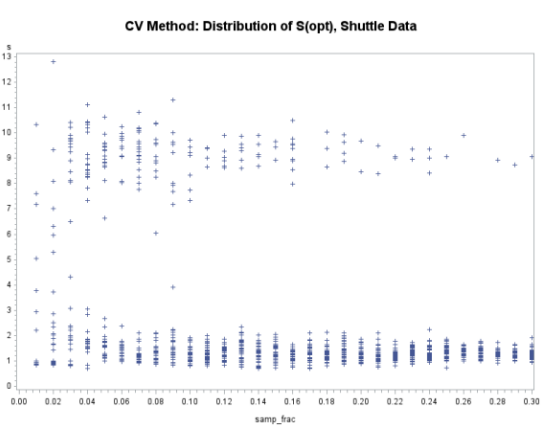

(a)  Mean and variance of $s_{opt}$ against sample size          (b)  Distribution of $s_{opt}$ against sample size

Figure 27. Sampling CV method for Shuttle data.

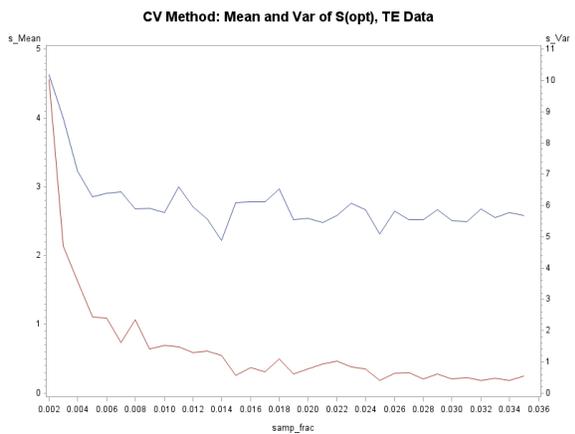 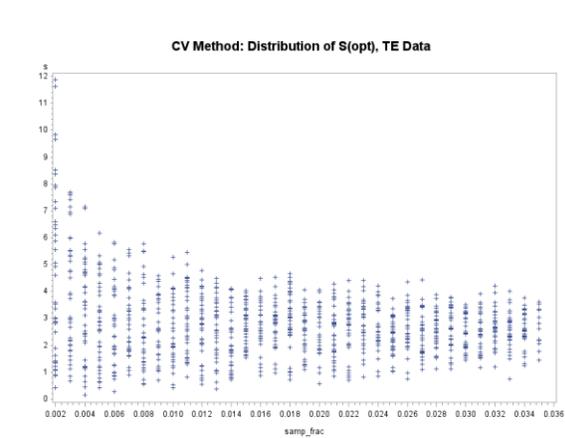

(a)  Mean and variance of $s_{opt}$ against sample size          (b)  Distribution of $s_{opt}$ against sample size

Figure 28. Sampling CV method for Tennessee Eastman data.

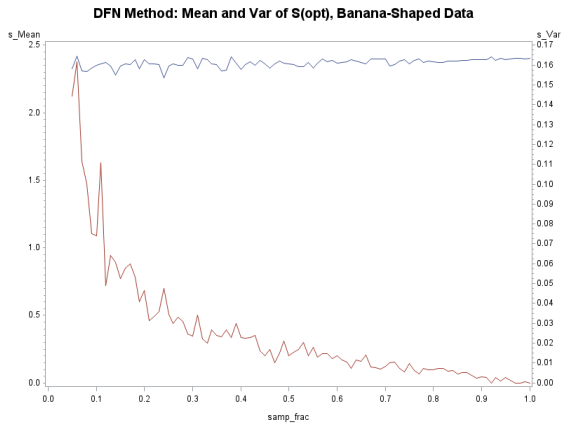
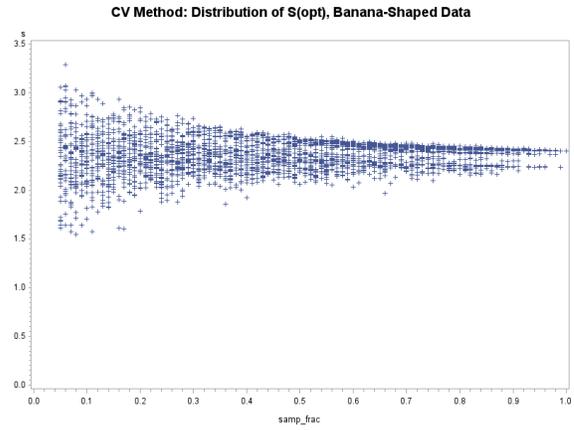

(a) Mean and variance of $s_{opt}$ against sample size

(b) Distribution of $s_{opt}$ against sample size

Figure 29. Sampling DFN method for banana-shaped data.

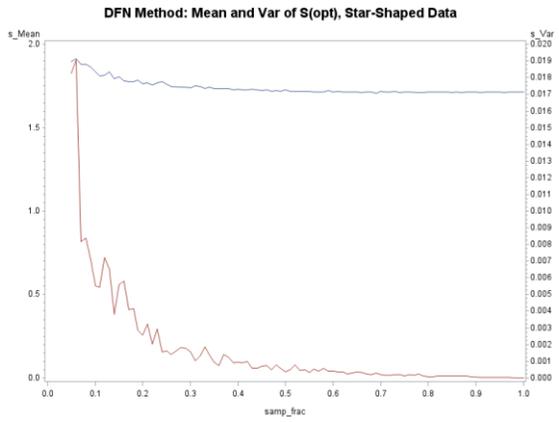
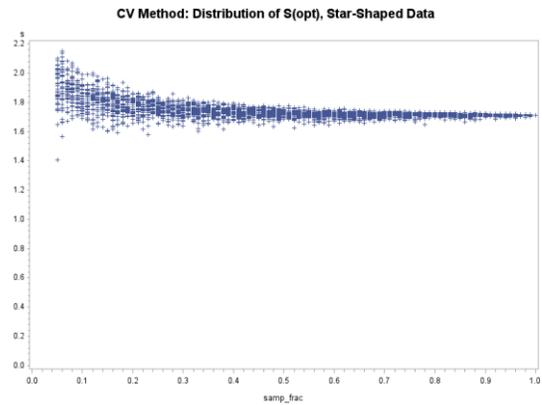

(a) Mean and variance of $s_{opt}$ against sample size

(b) Distribution of $s_{opt}$ against sample size

Figure 30. Sampling DFN method for star data.

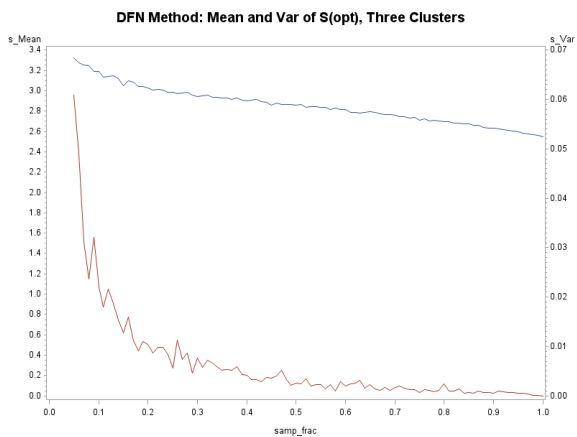
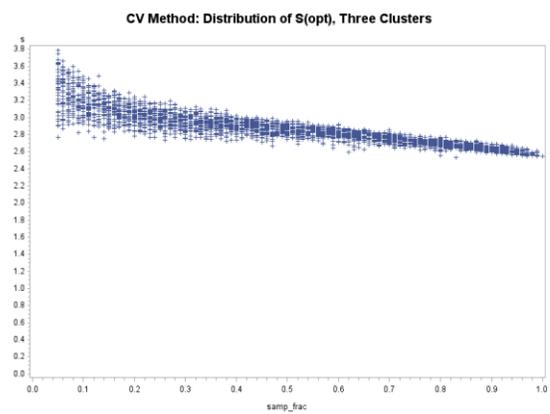

(a) Mean and variance of $s_{opt}$ against sample size

(b) Distribution of $s_{opt}$ against sample size

Figure 31. Sampling DFN method for three-cluster data.

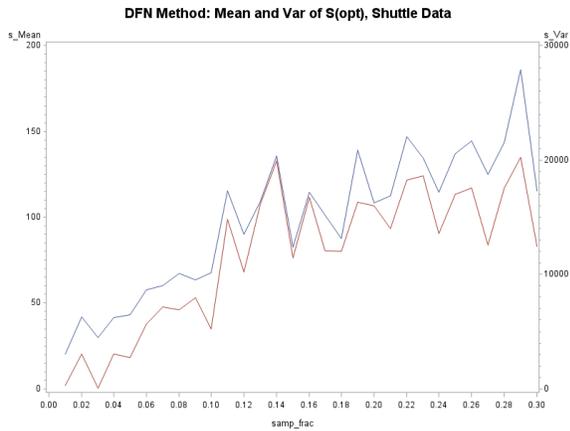

(a) Mean and variance of $s_{opt}$ against sample size

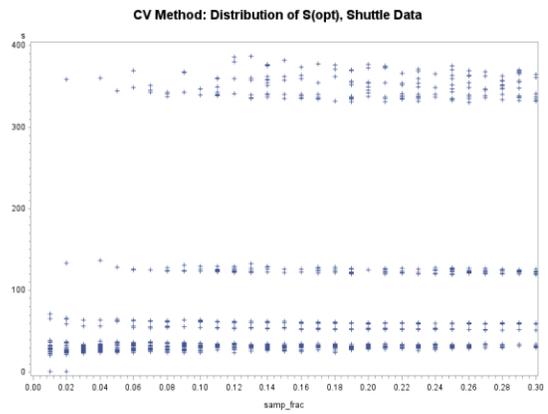

(b) Distribution of $s_{opt}$ against sample size

Figure 32. Sampling DFN method for Shuttle data.

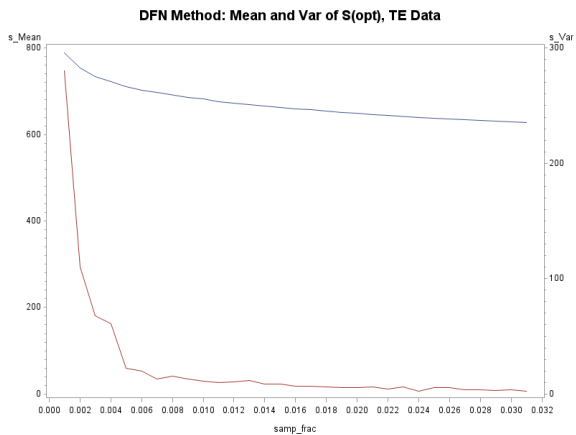

(a) Mean and variance of $s_{opt}$ against sample size

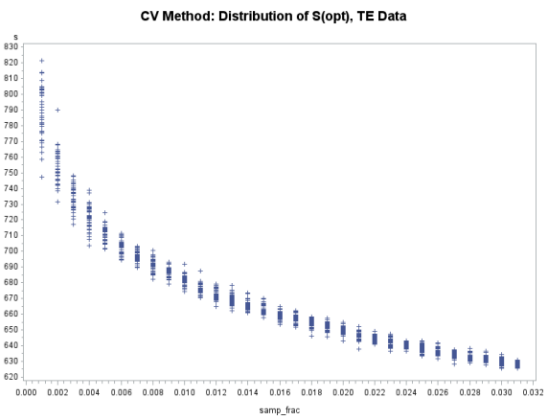

(b) Distribution of $s_{opt}$ against sample size

Figure 33. Sampling DFN method for Tennessee Eastman data.

## 5. Conclusions

This paper proposes an extension to the peak method that enables you to select an optimal value of the Gaussian kernel bandwidth parameter $s$ for SVDD training when training data sets are large. The proposed extension, called the sampling peak method, produces a value of $s$ that provides a good-quality boundary that conforms to the geometry of the training data set. Sampling SVDD for a range of values of $s$ is applied, and approximation to the optimal $s$ is obtained at the first maximum of the first derivative of the optimal objective function. As sample size is increased, we observed convergence of an optimal $s$ to an overall optimal value. The good convergence is demonstrated for relatively small sample sizes of large multidimensional data. The proposed method is shown to perform better than the existing alternative methods, such as coefficient of variation or distance to the farthest neighbor.

One potential area for further research is finding an optimal number of knots in the smoothing step in order to automate the selection of optimal $s$.

## Appendix A. Mathematical Formulation of SVDD

In the primal form, the SVDD model for normal data description builds a minimum-radius hypersphere around the data, with an objective function,

$$\min_{R,a} R^2 + C \sum_{i=1}^{n} \xi_i \tag{A.1}$$

subject to

$$\|x_i - a\|^2 \leq R^2 + \xi_i, \ \forall \ i = 1, \dots, n \tag{A.2}$$
$$\xi_i \geq 0, \forall \ i = 1, \dots, n \tag{A.3}$$

where

$x_i \in \mathbb{R}^m, i = 1, \dots, n$ is the training data

$R$ is the radius ($R \in \mathbb{R}$), which is a decision variable

$\xi_i$ is the slack for each observation, ($\xi_i \in \mathbb{R}$)

$a$ is the center ($a \in \mathbb{R}^m$), which is a decision variable

$C \left( = \frac{1}{n \cdot f} \right)$ is the penalty constant, the tradeoff between the volume of the hypersphere and the outliers

$f$ is the expected outlier fraction

The dual formulation of SVDD model is obtained using the Lagrange multipliers. The objective function is

$$\max_{\alpha_i} \sum_{i=1}^{n} \alpha_i \left( x_i \cdot x_i \right) - \sum_{i,j} \alpha_i \alpha_j \left( x_i \cdot x_j \right) \tag{A.4}$$

subject to

$$\sum_{i=1}^{n} \alpha_i = 1 \tag{A.5}$$
$$0 \leq \alpha_i \leq C, \ \forall \ i = 1, \dots, n \tag{A.6}$$

where $\alpha_i$ is the Lagrange multiplier ($\alpha_i \in \mathbb{R}$) and $\left( x_i \cdot x_j \right)$ is the inner product.

Solving (A.4) subject to (A.5) and (A.6) results in a spherical boundary around the data. If the expected data distribution is not spherical, then extra space might be included within the boundary, leading to false positives when scoring. It is therefore desirable to have a compact boundary around the data that would conform to the geometric shape of the single-class training data. This is possible if the inner product $\left( x_i \cdot x_j \right)$ is replaced with a suitable kernel function $K\left( x_i, x_j \right)$. This paper uses the Gaussian kernel function,

$$K\left( x_i, x_j \right) = exp \frac{-\|x_i - x_j\|^2}{2s^2} \tag{A.7}$$

where $s$ is the Gaussian bandwidth parameter.

The modified formulation of SVDD with the kernel function includes the following objective function,

$$\max_{\alpha_i} \sum_{i=1}^{n} \alpha_i K(x_i, x_i) - \sum_{i,j} \alpha_i \alpha_j K(x_i, x_j) \qquad (A.8)$$

subject to

$$\sum_{i=1}^{n} \alpha_i = 1, \qquad (A.9)$$

$$0 \leq \alpha_i \leq C, \; \forall \; i = 1, \dots, n \qquad (A.10)$$

The solution to the objective function (A.8) subject to (A.9) and (A.10) provides the values of $\alpha_i$. These values are used to compute a threshold $R^2$, which is calculated as

$$R^2 = K(x_k, x_k) - 2 \sum_{i=1}^{n} \alpha_i K(x_i, x_k) + \sum_{i,j} \alpha_i \alpha_j K(x_i, x_j) \qquad \forall x_i \in SV_{<c}, \qquad (A.11)$$

where $SV_{<c}$ is the set of support vectors for which $\alpha_k < C$.

For each observation $z$ in the scoring data set, the distance $D^2(z)$ is calculated as follows:

$$D^2(z) = K(z, z) - 2 \sum_{i=1}^{n} \alpha_i K(x_i, z) + \sum_{i,j} \alpha_i \alpha_j K(x_i, x_j) \qquad (A.12)$$

Any point from the scoring data set for which $D^2(z) > R^2$ is identified as an outlier.

# References

1. Kakde, Deovrat, Arin Chaudhuri, Seunghyun Kong, Maria Jahja, Hansi Jiang, and Jorge Silva. Peak Criterion for Choosing Gaussian Kernel Bandwidth in Support Vector Data Description. *arXiv preprint arXiv:1602.05257* (2016).

2. Chaudhuri, Arin, Deovrat Kakde, Maria Jahja, Wei Xiao, Hansi Jiang, Seunghyun Kong, and Sergiy Peredriy. Sampling Method for Fast Training of Support Vector Data Description. *arXiv preprint arXiv:1606.05382* (2016).

3. M. Lichman. UCI machine learning repository, 2013.

4. N. Laurence Ricker. Tennessee Eastman challenge archive, matlab 7.x code, 2002.

5. Paul F Evangelista, Mark J Embrechts, and Boleslaw K Szymanski. Some properties of the Gaussian kernel for one class learning. *In Artificial Neural Networks–ICANN 2007*, pages 269–278. Springer, 2007.

6. Yingchao Xiao, Huangang Wang, Lin Zhang, and Wenli Xu. Two methods of selecting Gaussian kernel parameters for one-class SVM and their application to fault detection. *Knowledge-Based Systems*, 59:75–84, 2014.

7. Tax, David MJ and Duin, Robert PW. Support vector data description. *Machine learning*, 54(1):45–66, 2004.

8. Thuntee Sukchotrat, Seoung Bum Kim, and Fugee Tsung. One-class classification-based control charts for multivariate process monitoring. *IIE transactions*, 42(2):107–120, 2009.

9. Achmad Widodo and Bo-Suk Yang. Support vector machine in machine condition monitoring and fault diagnosis. *Mechanical Systems and Signal Processing*, 21(6):2560–2574, 2007.

10. Alexander Ypma, David MJ Tax, and Robert PW Duin. Robust machine fault detection with independent component analysis and support vector data description. In Neural Networks for Signal Processing IX, 1999. *Proceedings of the 1999 IEEE Signal Processing Society Workshop*, pages 67–76. IEEE, 1999.

11. Carolina Sanchez-Hernandez, Doreen S Boyd, and Giles M Foody. One-class classification for mapping a specific land-cover class: Svdd classification of fenland. *Geoscience and Remote Sensing, IEEE Transactions on*, 45(4):1061–1073, 2007.